\documentclass[10pt,twocolumn,letterpaper]{article}

\usepackage{cvpr}              

\input{preamble}

\definecolor{cvprblue}{rgb}{0.21,0.49,0.74}
\usepackage[pagebackref,breaklinks,colorlinks,citecolor=cvprblue]{hyperref}

\def\paperID{5} 
\def\confName{CVPR}
\def\confYear{2024}

\title{Leveraging Image Matching Toward End-to-End Relative Camera Pose Regression}

\author{Fadi Khatib* \hspace{1cm}  
Yuval Margalit*  \hspace{1cm}
Meirav Galun \hspace{1cm}
Ronen Basri \\[0.1cm] Weizmann Institute of Science
}

\begin{document}
\maketitle

{\let\thefootnote\relax\footnotetext{*Equal contributors.
}}
{\let\thefootnote\relax\footnotetext{Project webpage:\\ \url{https://fadikhatib.github.io/GRelPose}
}}

\begin{abstract}
This paper proposes a generalizable, end-to-end deep learning-based method for relative pose regression between two images. Given two images of the same scene captured from different viewpoints, our method predicts the relative rotation and translation (including direction and scale) between the two respective cameras.
Inspired by the classical pipeline, our method leverages Image Matching
(IM) as a pre-trained task for relative pose regression. 
Specifically, we use LoFTR, an architecture that utilizes an attention-based network pre-trained on Scannet, to extract semi-dense feature maps, which are then warped and fed into a pose regression network. Notably, we use a loss function that utilizes separate terms to account for the translation direction and scale. We believe such a separation is important because translation direction is determined by point correspondences while the scale is inferred from prior on shape sizes. Our ablations further support this choice. We evaluate our method on several datasets and show that it outperforms previous end-to-end methods. The method also generalizes well to unseen datasets.
\end{abstract}    
\section{Introduction}
\label{sec:intro}
Computing the relative pose (position and orientation) between two cameras given an image pair is a fundamental computer vision task and an essential step in multiview structure from motion (SfM) pipelines \cite{ozyecsil2017survey, snavely2006photo,schoenberger2016sfm, wu2011visualsfm, wu2013towards}, with applications to autonomous robot navigation and to virtual and augmented reality. Classical techniques rely on keypoint extraction and matching using descriptors such as SIFT \cite{lowe2004distinctive,bay2006surf,rublee2011orb,calonder2010brief,rosten2006machine}, followed by robust estimation of the essential matrix using, e.g., RANSAC \cite{fischler1981random}. These methods can recover the relative pose very accurately. However, (1) they only recover five of the six degrees of freedom of the relative pose; the distance between the two camera positions cannot be recovered from point correspondences; (2) RANSAC generally requires many iterations (faster versions were recently proposed, e.g., MAGSAC++ \cite{barath2020magsac++}), and (3) classical methods are sensitive to matching errors due to the presence of repetitive structures or lighting differences.

\begin{figure*}{}
    \centering
    \includegraphics[width=0.85\textwidth]{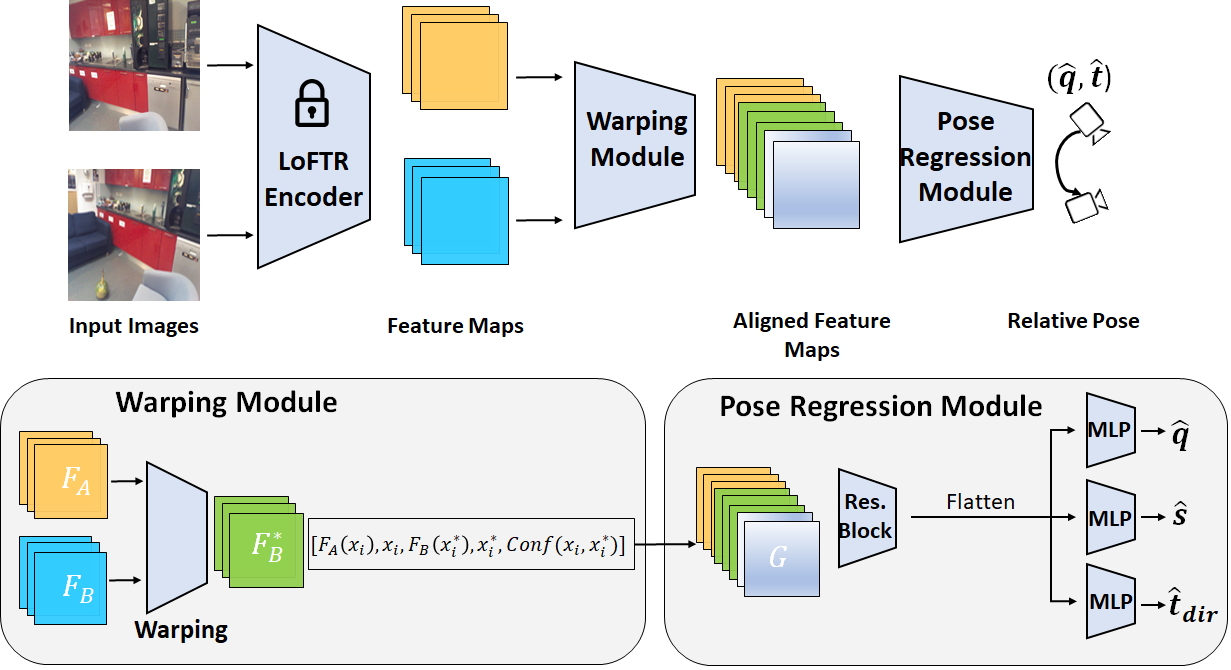}
    \caption{\textbf{Network architecture}. Our network is composed of three modules (top), LoFTR which given an image pair produces a coarse grid of features, a warping module, and a pose regression module. The bottom figure shows a detailed description of the warping and pose regression modules.} \label{fig:network_architecture}
\end{figure*}

To overcome these drawbacks, a number of recent papers have explored the use of deep neural networks for relative pose regression (RPR). While some methods replaced only parts of the pipeline (e.g., the keypoint extraction and matching, see Section~\ref{sec:related work} for discussion), our focus is on end-to-end RPR network solutions. These new RPR methods use appearance and pose priors to guide the matching of point features and to select the likely relative pose given an input image pair. In particular, deep methods can recover the distance between two cameras by exploiting priors on the size of the observed objects. Nevertheless, current deep methods are not yet as accurate as the classical approach in many settings \cite{barroso2023fsnet,kendall2015posenet,kendall2017geometric}. Moreover, existing methods still struggle with cross-scene/dataset generalization. 

Below we propose a three-step, deep network architecture for relative pose regression (see architecture in Figure~\ref{fig:network_architecture}). Similar to the classical techniques, our network proceeds by first computing feature descriptors for each image, then matching these features across the two images and finally regressing the motion parameters. For the first step we compute for each image a semi-dense feature map using the pre-trained LoFTR architecture \cite{sun2021loftr}. The second step matches and warps the most similar feature in the second image to each feature in the first image. The final step uses a residual network to compute the pose from the warped features.

We only train the last step in this pipeline. For training, we use a modified loss, compared to previous work \cite{arnold2022map,rockwell20228,laskar2017camera}. Noticing that translation direction is determined by the point matches, whereas the scale of translation is determined by a prior on the object size, we separate these in the loss: we use a cosine similarity term to train for direction and an $\ell_1$ term to train for scale. An additional term is used to train for orientation. We balance these terms simply and uniformly with the constant 1 and explore a similar, three-head architecture that requires no balancing of terms.

The main contributions of our work are as follows. (1) We leverage IM as a pre-training task for relative pose regression, offering a novel perspective on end-to-end relative pose estimation.
(2) Unlike the previous works, our loss treats separately the direction and scale of the camera position vector, using the cosine similarity for the directional loss and an $\ell_1$ loss for the scale. (3) We use hard matching and warping, instead of soft matching and warping as in \cite{arnold2022map}, we show empirically the advantage of our choice. (4) We validate that our pre-trained IM backbone generates effective feature representations for RPRs, highlighting also the contribution of interleaved self- and cross-attention modules in capturing feature similarities across image pairs.
(5) We test our method on various datasets, including cases in which the training and test datasets differ (e.g., training on indoor scenes and testing on outdoor scenes). Our method achieves superior results to other end-to-end RPR networks in nearly all experiments.
(6) Our method has made significant advancement in closing the performance gap between RPRs and feature matching methods, while maintaining substantially faster inference time.


\section{Related Work}
\label{sec:related work}

\noindent\textbf{Relative Pose Estimation}.
The pose between two images with
known intrinsic parameters can be recovered by decomposing the essential matrix \cite{hartley2003multiple},
yielding a relative rotation and a scaleless translation direction vector. The essential
matrix is classically estimated by matching local features, such as \cite{lowe2004distinctive,bay2006surf,rublee2011orb,calonder2010brief,rosten2006machine}, followed by a numerical scheme, e.g., the 5-point solver \cite{nister2004efficient} inside robust estimator loop, namely RANSAC \cite{fischler1981random} and others \cite{barath2019magsac,barath2020magsac++,raguram2008comparative}. Despite the success of this classical approach, it lacks the ability to utilize priors, for example, to infer the unknown distance between the two cameras. These shortcomings have led to the introduction of deep-based approaches, which we review next.

\noindent\textbf{Deep feature extraction and matching}. Several studies have utilized deep networks to detect and match point features in image pairs \cite{revaud2019r2d2,dusmanu2019d2,bhowmik2020reinforced,sarlin20superglue,tyszkiewicz2020disk,yi2016lift,truong2020glu,truong2020gocor,truong2021learning,sun2021loftr,jiang2021cotr}. Among these, LIFT \cite{yi2016lift} is an early approach that additionally assigns an orientation to each feature. Superglue \cite{sarlin20superglue} uses a Graph Neural Network (GNN) to match two sets of interest points. Lightglue \cite{lindenberger2023lightglue} uses self and cross attention alongside key-point pruning and early adaptive depth mechanism to improve on Superglue's inference speed and accuracy, utilizing keypoints and descriptors extracted, for example, with SuperPoint\cite{detone2018superpoint} . A number of works utilize neural networks to construct a differentiable matching module (providing also matching confidence) \cite{truong2020glu,truong2020gocor,truong2021learning}. The recent LoFTR \cite{sun2021loftr} and COTR 
\cite{jiang2021cotr} utilize self- and cross-attention layers along with multiscale analysis to produce semi-dense, near-pixel-wise correspondences. The relative pose is then obtained by applying RANSAC to the extracted correspondences. Those methods were shown to produce accurate correspondences even in low-texture areas, demonstrating good generalization ability across scenes and datasets.  Still, those methods are not end-to-end approaches, and the relative position is recovered up to a scale factor.

\noindent\textbf{Deep RANSAC}. Recent works sought to construct a deep network architecture that can robustly estimate the essential matrix from point matches. Specifically, DSAC~\cite{brachmann2017dsac} replaces RANSAC's hypothesis selection with a robust average of hypotheses, resulting in a differentiable counterpart of RANSAC. \cite{brachmann2023ace} suggests using a fixed backbone for the DSAC\textsuperscript{*}  network and training only a scene-specific regression head. Alongside a novel training loss, it resulted in faster convergence and better accuracy. 
\cite{choy2020high} utilizes convolutions in high dimension to recover the essential matrix robustly.  \cite{ranftl2018deep,brachmann2019expert,brachmann2019neural} propose a deep-based scheme to distinguish between inliers and outliers, given putative correspondences, for the tasks of fundamental matrix estimation and camera re-localization.

\noindent\textbf{End-to-end relative pose regression (RPR)}.
A number of deep learning methods attempt to predict the relative pose directly from input image pairs, circumventing the need to detect 2D putative matches explicitly.
Early works \cite{melekhov2017relative,laskar2017camera} used ResNet34 pre-trained for classification on the Imagenet dataset to produce holistic feature embeddings that describe the two input images. An MLP module was then applied to these feature vectors to regress the relative position, with its magnitude, as well as the relative orientation. Subsequent works \cite{abouelnaga2021distillpose, zhou2020learn, winkelbauer2021learning, en2018rpnet,balntas2018relocnet} utilized a similar approach using modified loss functions and architectures. All of these methods have limited cross-scene generalization ability.
Recent works \cite{zhang2022relpose, lin2023relpose++} have utilized energy-based relative pose models for multi-view pose estimation, demonstrating their efficacy in sparse object-centric scenarios exemplified by the CO3D dataset \cite{reizenstein2021common}. Their applicability to more general scenes, however, remains a challenge.

 Recently, \cite{rockwell20228} advocated the use of Vision Transformer (ViT) feature maps \cite{dosovitskiy2020image, vaswani2017attention} for relative pose regression. They further incorporated quadratic positional encoding to produce an architecture that mimics the Eight-Point Algorithm for essential matrix recovery \cite{hartley1997defense}. Lastly, The Map-free benchmark~\cite{arnold2022map} suggests baselines with different output parametrizations for RPR, using a residual UNet (ResUNet) architecture to produce a feature map for each image. A soft feature matching was obtained by computing a 4D correlation volume followed by softmax. This soft matching was then used to warp the features of one image into their corresponding locations in the other image. A regression module was finally used to estimate the relative pose.

\section{Method}
\label{sec:method}

Our goal, given two images of a scene captured from different viewpoints, is to predict the relative position and orientation between the two respective cameras, including the distance between their centers. Similar to recent approaches to end-to-end RPR networks \cite{arnold2022map,rockwell20228}, we use a three-module architecture (see Fig.~\ref{fig:network_architecture}), a frozen feature extraction module, a matching and warping module, and a pose regression head. Our architecture improves over previous work by relying on the powerful LoFTR IM architecture~\cite{sun2021loftr} to produce a semi-dense feature map, warping the feature maps by utilizing hard matches between the produced features, followed by a camera motion regressor, which is trained by a loss in which translation direction and scale are separated into different terms. These modifications allow us to significantly improve accuracy and achieve cross-scene and
cross-dataset generalization. 
Below we describe the three modules and our loss function.

\subsection{Feature extraction}

For feature extraction we use the coarse module of LoFTR \cite{sun2021loftr}, which is pre-trained for matching on the Scannet dataset \cite{dai2017scannet}, depicting indoor scenes. We keep the weights of this module frozen throughout the training of our network. The LoFTR architecture is applied to both images in parallel. Denote the input images by $\Ima$ and $\Imb$. Each image is first fed to a feature pyramid network \cite{lin2017feature}, producing a pixel-wise feature map at the desired resolution. Next, a positional encoding is summed to these maps. The obtained maps are then fed into a sequence of interleaved self- and cross-attention layers, producing the final feature maps, $F_A(\x_i),F_B(\x_j) \in \Real^C$, where $C$ denotes the number of feature channels, $1 \le i,j \le n$, $n=WH/64$, and $W$ and $H$ denote the width and height of the input images in their original resolution. Note that, unlike in~\cite{arnold2022map}, the cross-attention steps modify the feature maps to account for the features seen in the other image, potentially encouraging their matching.

\subsection{Matching and warping}

Our next aim is to identify matching points. Given the coarse feature maps $F_A(\x_i)$ and $F_B(\x_j)$ (expressed as column vectors), we select for each point $\x_i$ in $\Ima$ a corresponding point $\x_i^*$ in $\Imb$ by maximizing the correlation between their respective feature vectors, $F_A(\x_i)$ and $F_B(\x_i^*)$. Specifically, the coordinate $\x_i^*$ in $\Imb$ is determined by
\begin{equation} \label{eq:warping_eq}
    \x_i^* = \argmax_j (F_A^T(\x_i) F_B(\x_j)).
\end{equation}
Additionally, we use the computed correlations $F_A^TF_B$ to calculate a confidence score by applying softmax over each collection of correlations for point $\x_i$ in the first image, namely,

\begin{equation}
    \conf(\x_i,\x_i^*) = \max(\underset{j}{\mathrm{softmax}} (F_A^T(\x_i) F_B(\x_j))).
\end{equation}

Next, we concatenate corresponding features, append an encoding of their positions in the two images, and the confidence score, obtaining a feature map of correspondences, namely,
\begin{equation}
    G(\x_i) = [F_A^T(\x_i),\x_i, F_B^T(\x_i^*), \x_i^*,\conf(\x_i,\x_i^*)].
\end{equation}

\begin{figure}[t]   
 
     \includegraphics[width=0.9\columnwidth]{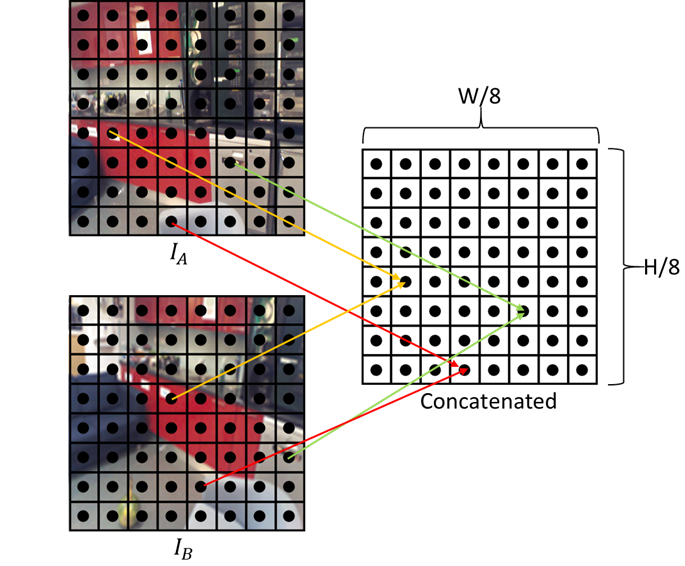}
     \caption{\textbf{The warped feature map.} Given a feature vector
from Image A, we compute the dot product with all feature vectors
in Image B, and we use it to warp the features in $F_B$ their corresponding features in $F_A$, see \eqref{eq:warping_eq}. }
\label{fig:aligning}
 \end{figure}

Note that in this feature map, the features in $F_B$ are warped to their corresponding features in $F_A$ (see Fig. \ref{fig:aligning}).

\subsection{Pose Regression module}

Given the warped feature map, $G(\x_i)$, we apply a sequence of residual blocks followed by an MLP that regresses the different components of the relative pose,  $(\hat{\q},\hat{\tr}_{dir},\hat{s}) \in \Real^8$, where $\hat\q \in \Real^4$ denotes the predicted rotation represented as a quaternion,  $\hat{\tr}_{dir} \in \Real^3$ denotes the signed direction of the translation, and $\hat{s}$ is a scalar, denoting the scale of the translation.
The quaternion $\hat\q$ and translation direction $\hat{\tr}_{dir}$ are normalized to unit magnitude to respectively depict a rotation and an orientation.

\subsection{Loss Function}

We use supervised learning to train the pose regression module. Given image pairs with ground truth pose specified by a rotation quaternion $\q$ and position $\tr$, we let
\begin{equation}
\mathcal{L}_{\q}(\Ima,\Imb) = \left\|\frac{\hat\q}{\|\hat\q\|}-\q\right\|_{1},
\label{eq:loss_rot}
\end{equation}
\begin{equation}
\mathcal{L}_{\tr_{dir}}(\Ima,\Imb)=1-\frac{\inner{\hat{\tr}_{dir}}{\tr}}{\|\hat\tr_{dir}\|\|\tr\|},
\label{eq:loss_trans}
\end{equation}
and
\begin{equation}
\mathcal{L}_{s}(\Ima,\Imb)=|\hat{s}-\|\tr\||,
\label{eq:loss_normtrans}
\end{equation}
where $\mathcal{L}_{\q}$ measures the rotation error  $\mathcal{L}_{\tr_{dir}}$ depicts the direction error of the translation, and $\mathcal{L}_{s}$ depicts the metric error in the translation. A quaternion $\q$ has the form $(\cos(\theta/2),\vv\sin(\theta/2))$, where $\theta$ represents a rotation angle and $\vv$ represents an axis of rotation.
This means that $\pm \q$ represents the same rotation; therefore, to avoid ambiguity, we set its first entry in the ground truth annotation to be positive.

As \cite{kendall2017geometric} shows, balancing rotation and translation terms can be challenging. Here we simply assign equal weights to each term in the loss, i.e., 
\begin{align}
\mathcal{L}(\Ima,\Imb) & = \mathcal{L}_{\q}(\Ima,\Imb) +  \mathcal{L}_{\tr_{dir}}(\Ima,\Imb) + \mathcal{L}_{s}(\Ima,\Imb).
\label{eq:loss}
\end{align}
Our experiments in Sec.~\ref{sec:experiments} indicate that this choice is suitable for indoor relative pose estimation. 

\section{Experiments}
\label{sec:experiments}

\begin{table*}[!ht]
    \centering
    \begin{tabular}{|l|c|c|c|c|}    
    \hline
        Scene/Method & RelPoseNet\cite{laskar2017camera} & 8Point\cite{rockwell20228} & RPR [$q+t$]\cite{arnold2022map}  & Ours  \\ \hline
        \hline
        Chess & 26.0/51.5/8.14 & 12.3/27.2/\textbf{3.38} & 15.4/41.6/5.84  &\textbf{9.1}/\textbf{19.9}/3.84 \\ \hline
        Fire & 38.2/70.2/12.3 & 27.0/46.2/7.82 & 33.4/69.5/12.2  &\textbf{13.6}/\textbf{22.8}/\textbf{5.09}\\ \hline
        Heads & 23.3/62.0/12.4 & 17.2/53.0/10.4 & 19.1/58.5/12.5  &\textbf{9.1}/\textbf{23.6}/\textbf{5.77} \\ \hline
        Office & 30.7/53.0/9.04 & 18.1/29.6/4.06 & 23.7/44.6/7.32  &\textbf{11.2}/\textbf{16.6}/\textbf{3.79} \\ \hline
        Pumpkin & 35.1/44.5/9.36 & 18.7/26.1/3.68 & 26.9/36.9/6.65  &\textbf{13.9}/\textbf{19.7}/\textbf{3.66} \\ \hline
        Red Kitchen & 40.7/70.7/9.63 & 21.0/39.3/4.15 & 26.7/51.3/7.09 & \textbf{12.7}/\textbf{19.1}/\textbf{3.62}\\ \hline
        Stairs & 36.7/77.3/12.2 &\textbf{26.3}/50.4\textbf{/5.02}  & 31.3/62.9/10.2  &26.5/\textbf{44.2}/6.05\\ \hline
        Average & 32.9/61.3/10.4 & 21.4/38.8/5.49 & 26.8/59.8/8.83  &\textbf{13.7}/\textbf{23.7}/\textbf{4.55}\\ \hline
    \end{tabular}
    \caption{\textbf{7Scenes experiment}. The table shows the result of training and testing on the full 7Scenes dataset with the train and test split specified in the dataset. Each entry shows position error in centimeters (left),  position angular error in degrees (middle), and orientation error in degrees (right).}
    \label{tab:7scenes}
\end{table*}

\subsection{Datasets}

In each of the following experiments, we train our network on either the 7Scenes \cite{glocker2013real} or Scannet \cite{dai2017scannet} indoor datasets or on the Map-Free \cite{arnold2022map} outdoor dataset. We evaluate our method on the 7Scenes and Scannet for the indoor case and for the outdoor on the Map-Free and Cambridge landmarks~\cite{kendall2015posenet} datasets. See the supplementary material for how the GT relative pose is computed.

\noindent\textbf{7Scenes \cite{glocker2013real}}. The Microsoft 7Scenes dataset is a widely used indoor dataset for visual relocalization that contains RGB-D images covering seven different indoor locations (rooms) scanned with KinectFusion. Each scene is provided with training and testing image sequences. We use the dataset as provided in \cite{laskar2017camera, melekhov2017relative}, consisting of roughly 40K pairs of images and 80K test pairs.

\noindent\textbf{Cambridge Landmarks \cite{kendall2015posenet}}. An urban localization dataset capturing four outdoor scenes in Cambridge, UK. The ground truth poses are determined using a standard SfM method (with positions recovered up to an unknown global scale). The images were captured at different times, affected by different lighting and weather condition and containing moving objects including vehicles and pedestrians. 

\noindent\textbf{Scannet~\cite{dai2017scannet}} Scannet is a huge indoor dataset consisting of 1613 monocular sequences with ground truth poses and depth maps. This
dataset contains image pairs with wide baselines and extensive texture-less regions. We use the same train/test split used by LoFTR~\cite{sun2021loftr}, where we sample 200 pairs per scene. This results in 299,200 training pairs (that change in training at each epoch).

\noindent\textbf{Map-Free~\cite{arnold2022map}} The Map-free benchmark is a large outdoor dataset consisting of 655 places of interest including sculptures, murals and fountains collected worldwide, such that a place can be well-captured by a single image. Each space comes with a single reference image that serves as the relocalization anchor; all the others are query images with metric poses. This is an extremely challenging setting with limited overlap between image pairs, illumination changes, and different camera intrinsics between sequences as can be seen in Fig~\ref{fig:mapfree}. The training set consists of 460 training scenes, 65 validation scenes, and 130 testing scenes.

\subsection{Baselines}
\begin{table*}[!h]
    \centering
    \begin{tabular}{|l|c|c|c|c|}
        \hline
        Scene/Method & RelPoseNet\cite{laskar2017camera} & 8Point\cite{rockwell20228} & RPR [$q+t$]\cite{arnold2022map} & Ours \\ \hline
        \hline
        Chess & 22.8/48.2/7.13 & 27.9/79.6/8.77 & 27.4/83.5/10.2 &  \textbf{18.25}/\textbf{36.72}/\textbf{5.48}\\ \hline
        Fire & 37.4/67.1/19.4 & 40.6/93.6/12.2 & 34.0/75.0/13.0 & \textbf{19.94}/\textbf{32.2}/\textbf{7.78}
\\ \hline
        Heads & 26.5/69.4/12.8 & 28.0/79.8/10.7 & 25.8/83.4/14.0 & \textbf{15.23}/\textbf{25.92}/\textbf{6.36}
\\ \hline
        Office & 28.4/54.5/9.63 & 35.2/81.1/8.84 & 34.8/79.5/11.2 & \textbf{18.06}/\textbf{27.14}/\textbf{4.99}
\\ \hline
        Pumpkin & 34.9/47.4/10.3 & 42.3/88.5/9.77 & 36.9/69.4/10.9 & \textbf{23.66}/\textbf{35.29}/\textbf{5.3}
\\ \hline
        Red Kitchen & 39.0/69.1/9.73 & 32.7/74.7/10.3 & 34.2/81.2/11.7 & \textbf{18.22}/\textbf{26.37}/\textbf{5.15}
\\ \hline
        Stairs & 45.6/67.2/12.7 & 36.8/79.3/8.98  & 37.0/96.6/11.7 & \textbf{28.6}/\textbf{47.3}/\textbf{8.04}\\ \hline
        Average & 33.5/60.4/11.7 & 34.8/82.4/9.94 & 32.9/81.2/11.8 & \textbf{20.2}/\textbf{33.0}/\textbf{6.16}\\ \hline
    \end{tabular}
    \caption{\textbf{Six Scenes experiment}. The table shows the result of training on six scenes from the 7Scenes dataset and testing on the remaining scene. Each entry shows position error in centimeters (left), position angular error in degrees (middle), and orientation error in degrees (right).\\}
    \label{tab:six_scenes}
\end{table*}
We compare our approach to three recent end-to-end RPR methods described below.
For implementation and training details see the supplementary material.

\noindent\textbf{RelPoseNet} \cite{laskar2017camera}. This method uses ResNet34 classification network pre-trained on Imagenet dataset to extract a feature vector for each image. The two vectors are then concatenated and fed into a MLP that regresses the relative pose
.

\noindent\textbf{8-Point \cite{rockwell20228}}. This method produces, given an image, a feature embedding using a pre-trained Res-Net18 \cite{he2016deep} followed by ViT-tiny \cite{dosovitskiy2020image}, a bi-linear cross-attention layer (that considers, for each image, features from the other image), and a quadratic positional encoding. An MLP is finally used to regress the relative position and orientation.

\noindent\textbf{Map-free RPR}. This method is taken from the baselines suggested in \cite{arnold2022map}. It uses a residual UNet (ResUNet) architecture to produce a feature map for each image. A soft feature matching is obtained by computing a 4D correlation volume followed by softmax. This soft matching is then used to warp the features of one image into their corresponding locations in the other image. A regression module is finally used to estimate the relative pose. This method uses no pre-trained components.

We tested the Map-free model in three variants. For the first variant, denoted here as RPR [$q+t$], we retrained the network from scratch to output rotation quaternion and scaled translation parameters. For the second and third variants, we used the model pre-trained on the large Scannet dataset using the 3D-3D and Discretized output methods (denoted as RPR [3D-3D] and RPR disc. respectively) described in \cite{arnold2022map}.

\subsection{Metrics and Evaluation}

For each method, we measure the position error defined as the Euclidean distance between the predicted and ground truth camera centers.
\begin{equation}
    t_{euc\,err}(\tr,\hat{\tr}) = \Vert \hat{\tr} - \tr \Vert_2
\end{equation}
as well as the angle (in degrees) between the predicted and ground truth directions of the camera center vectors
\begin{equation}
    t_{ang\,err}(\tr,\hat{\tr}) = \frac{180}{\pi}\arccos\left(\left\vert \frac{\hat{\tr}^T \tr}{\|\hat\tr\|\|\tr\|}\right\vert\right).
\end{equation}
We measure orientation error by the angle between the predicted and ground truth orientations, represented in quaternions
\begin{equation}
    R_{err}(\q,\hat{\q}) = 2\arccos\left(\left\vert \frac{\hat\q^T \q}{\|\hat\q\|}\right\vert\right).
\end{equation}
For all measures, we report the median error over the different input pairs.
For experiments on Scannet, we also report the AUC for the translation and rotation angular errors.

\subsection{Results}

\textbf{7Scenes  $\rightarrow$ 7Scenes.} In our first experiment, we trained and tested all methods on the 7Scenes dataset
. The results are shown in Table~\ref{tab:7scenes}. Our method (marked as "Ours" in the tables) outperforms the other methods in almost all of the tested scenes, achieving a noticeable improvement in the translation (both Euclidean distance and angular distance) and a smaller advantage over 8Point~\cite{rockwell20228} in the rotation error. Overall, on average, we improve the position prediction by almost $8 \mathrm{cm}$, the position direction prediction by $15^\circ$, and the rotation prediction by  $1^\circ$.

\begin{table*}
\footnotesize
\centering

\begin{tabular}{|c|c|c|c|c|c|c|}
\hline
\multirow{2}{*}{ Method } &
\multicolumn{2}{|c|}{All Keyframes}&
\multicolumn{2}{|c|}{2 Keyframes}&
\multicolumn{2}{|c|}{1 Keyframe}\\
\cline{2-7} & Error\textdownarrow & Prec.($25 \mathrm{~cm}, 5^{\circ}$ )\textuparrow& Error\textdownarrow & Prec.($25 \mathrm{~cm}, 5^{\circ}$)\textuparrow& Error\textdownarrow & Prec.($25 \mathrm{~cm}, 5^{\circ}$)\textuparrow\\
\hline DSAC*(single scene)\cite{brachmann2021visual} & $3 \mathrm{~cm}, 1.1^{\circ}$ & 0.97 & $31 \mathrm{~cm}, 7.4^{\circ}$ & 0.41 & $77 \mathrm{~cm}, 25.5^{\circ}$ & 0.22 \\
\hline
\hline LoFTR\cite{sun2021loftr}+5Pt+Depth Scale & $\textbf{\textcolor{BurntOrange}{12}}\mathrm{~cm}, \textbf{\textcolor{BurntOrange}{1.7}}^{\circ}$ & \textbf{\textcolor{BurntOrange}{0.83}} & $29 \mathrm{~cm}, \textbf{\textcolor{BurntOrange}{2.8}}^{\circ}$ & 0.50 & $33 \mathrm{~cm}, 6.4^{\circ}$ & \textbf{\textcolor{BurntOrange}{0.38}} \\
\hline SG\cite{sarlin20superglue}+5Pt+Depth Scale & $13 \mathrm{~cm}, 1.8^{\circ}$ & 0.81 & $\textbf{\textcolor{BurntOrange}{26}} \mathrm{~cm}, 3.3^{\circ}$ & \textbf{\textcolor{BurntOrange}{0.51}} & $\textbf{\textcolor{BurntOrange}{30}} \mathrm{~cm}, \textbf{\textcolor{BurntOrange}{5.4}}^{\circ}$ & 0.37 \\
\hline SIFT\cite{lowe2004distinctive}+5Pt+Depth Scale & $16 \mathrm{~cm}, 2.5^{\circ}$ & 0.74 & $41 \mathrm{~cm}, 7.1^{\circ}$ & 0.37 & $60 \mathrm{~cm}, 23.7^{\circ}$ & 0.24 \\
\hline
\hline RPR [3D-3D]\cite{arnold2022map} & $16 \mathrm{~cm}, 4.5^{\circ}$ & 0.60 & $42 \mathrm{~cm}, 8.3^{\circ}$ & 0.18 & $66 \mathrm{~cm}, 12.8^{\circ}$ & 0.11 \\
\hline RPR $[3 \mathrm{D}-3 \mathrm{D}]$ f.t. hard \cite{arnold2022map}& $17 \mathrm{~cm}, 4.2^{\circ}$ & 0.61 & $37 \mathrm{~cm}, 7.3^{\circ}$ & 0.20 & $50 \mathrm{~cm}, \textbf{\textcolor{blue}{10.8}}^{\circ}$ & 0.13 \\
\hline RPR disc.\cite{arnold2022map} & $18 \mathrm{~cm}, 4.9^{\circ}$ & 0.53 & $51 \mathrm{~cm}, 10.0^{\circ}$ & 0.12 & $67 \mathrm{~cm}, 15.9^{\circ}$ & 0.07 \\
\hline Ours&$\textbf{\textcolor{blue}{11}} \mathrm{~cm}, \textbf{\textcolor{blue}{3.2}}^{\circ}$ & \textbf{\textcolor{blue}{0.73}} & $\textbf{\textcolor{blue}{35}} \mathrm{~cm}, \textbf{\textcolor{blue}{6.86}}^{\circ}$ & \textbf{\textcolor{blue}{0.27}} & $\textbf{\textcolor{blue}{46}} \mathrm{~cm}, 11.1^{\circ}$ & \textbf{\textcolor{blue}{0.16}}\\
\hline
\end{tabular}
\caption{\textbf{7scenes relocalization}. The table shows the result of training on Scannet and testing on the 7Scenes test set for relocalization using 1,2 or all the mapped reference frames. The errors are the average of the median error over all the scenes.
"RPR [3D-3D] f.t. hard" refers to the [3D-3D]  network in \cite{arnold2022map}, fine-tuned over Scannet image pairs with a small overlap. DSAC\textsuperscript{*} is a single scene absolute pose regression network.
For this experiment, we also report the precision (Prec.) at the position error of $25 \mathrm{~cm}$, and rotation error of $5^{\circ}$. We use \textbf{\textcolor{blue}{blue}} to indicate the best result among the RPR methods, and \textbf{\textcolor{BurntOrange}{orange}} to indicate the best result among the feature matching methods.}
\label{tab:7scenes_reloc}
\end{table*}

\noindent\textbf{Six scenes  $\rightarrow$ $7^{th}$ scene.}
The next experiment tests cross-scene generalization. For this experiment, we trained the methods on six of the seven scenes in the 7Scenes dataset and tested them on the remaining scene. We repeated this experiment seven times, each time holding out a different scene. Results are shown in Table \ref{tab:six_scenes}. Our method outperforms all other methods in this scenario as well. Notice in particular the advantage in the position angular error, as both 8Point and RPR [$q+t$] obtain a near random average median angular error (larger than 80\textsuperscript{\textdegree }). Overall, on average, relative to the best existing method, we improve the position prediction by $12 \mathrm{cm}$, the position direction prediction by $27^\circ$, and the rotation prediction by $3.7^\circ$.

\begin{table*}[!t]
\centering
\begin{tabular}{|c|c|c|c|c|c|c|c|c|}
\hline \multirow{2}{*}{ Method } &  \multicolumn{3}{|c|}{Trans. AUC \textuparrow}& \multicolumn{3}{|c|}{Rot. AUC \textuparrow}&  \multicolumn{2}{|c|}{Pose median errors\textdownarrow}\\ 
\cline{2-9}& @ $5^{\circ}$ & $@ 10^{\circ}$ & $@ 20^{\circ}$   & @ $5^{\circ}$ & $@ 10^{\circ}$ &$@ 20^{\circ}$  & Trans cm & Rotation\textdegree\\
\hline
LoFTR\cite{sun2021loftr}+5Pt+Depth Scale& \textbf{\textcolor{BurntOrange}{23.4}} & \textbf{\textcolor{BurntOrange}{42.4}} & \textbf{\textcolor{BurntOrange}{60.6}}  & \textbf{\textcolor{BurntOrange}{52.6}} & \textbf{\textcolor{BurntOrange}{69.1}} &\textbf{\textcolor{BurntOrange}{79.7}} & \textbf{\textcolor{BurntOrange}{13}}& \textbf{\textcolor{BurntOrange}{1.8}}\\
\hline
SG\cite{sarlin20superglue}+5Pt+Depth Scale& 18.2& 35.6& 54.7& 43.5& 62.0&75.4& 15& 2.4\\
\hline
SIFT\cite{lowe2004distinctive}+5Pt+Depth Scale & 6.5& 14.7& 26.3& 24.4& 36.3&47.1& 38& 8.0\\
\hline
\hline
RPR [3D-3D]\cite{arnold2022map}  & 4.5& 13.6& 31.8& 28.1& 48.6&65.9& 20& 4.1\\
\hline    RPR disc.\cite{arnold2022map}  & 3.6& 11.4& 29.7& 25.3& 44.8&62.4&22&4.7\\
\hline Ours  &\textbf{\textcolor{blue}{6.4}}&  \textbf{\textcolor{blue}{18.8}}& \textbf{\textcolor{blue}{40.0}}& \textbf{\textcolor{blue}{33.3}}& \textbf{\textcolor{blue}{54.0}}&\textbf{\textcolor{blue}{70.5}}& \textbf{\textcolor{blue}{15}}& \textbf{\textcolor{blue}{3.4}}\\
\hline
\end{tabular}
\caption{\textbf{Scannet experiment}. The table shows the result of training on Scannet and testing on the Scannet-1500 test set. We use \textbf{\textcolor{blue}{blue}} to indicate the best result among the RPR methods, and \textbf{\textcolor{BurntOrange}{orange}} to indicate the best result among the feature matching methods.}
\label{tab:scannet_test}
\end{table*}

\noindent\textbf{Scannet $\rightarrow$ 7Scenes.}
The third experiment tests cross-dataset generalization. Here all methods are trained on the Scannet dataset and tested on the 7Scenes test set for the relocalization task. In this task, we evaluate the methods for relocalization with a query image against either a single keyframe, two keyframes, or all keyframes. The single keyframe case is equivalent to the relative pose regression task. In the two/all keyframes case, we first select one of the two/all keyframes whose distance in the DenseVLAD~\cite{torii201524} feature space to the query image is smaller and then perform RPR on the query and the selected keyframe. The split is taken from \cite{arnold2022map}. Results are shown in Table \ref{tab:7scenes_reloc}.
Our method outperforms all other RPR methods, improving both translation (metric and angular) and rotation errors (with the exception of the finetuned RPR [3D-3D], which achieves a slight rotation advantage in the single keyframe case.). Nevertheless, while our method achieves significantly better performance than SIFT \cite{lowe2004distinctive}, its performance is still surpassed by learnable feature matching methods \cite{sun2021loftr,sarlin20superglue} combined with a depth estimation method \cite{liu2019planercnn}, for obtaining the metric scale from the depth as explained in \cite{arnold2022map}.

\begin{figure}
    \centering
    \includegraphics[width=0.9\columnwidth]{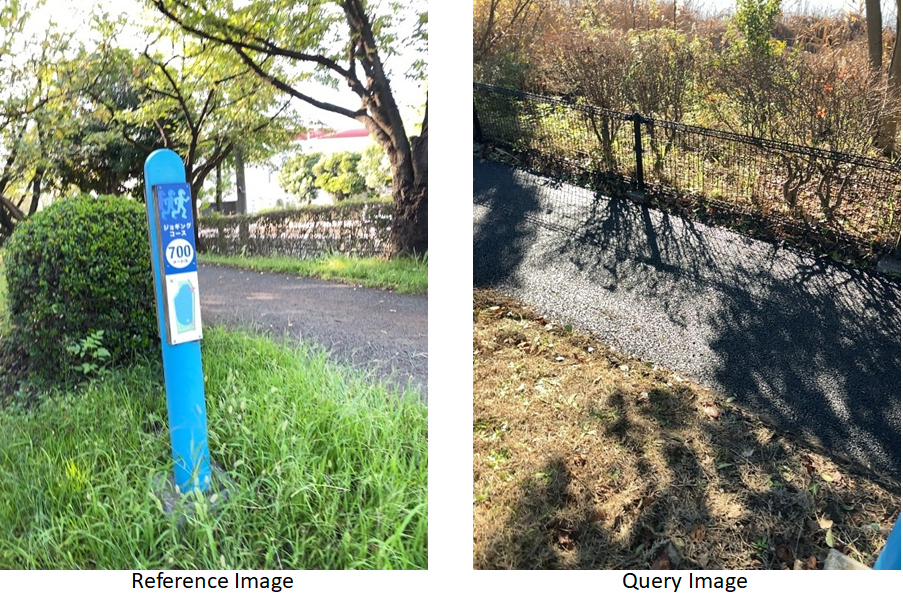}
    \caption{A pair of images from the Map-Free test set. There is little overlap and a significant change in the color intensities, probably due to a time difference between the images.}
    \label{fig:mapfree}
\end{figure} 

\noindent\textbf{ Scannet $\rightarrow$ Scannet}.
We also evaluate our methods on the Scannet-1500 test set for further cross-scene evaluation. As seen in Table~\ref{tab:scannet_test}, our method improves over the baselines with a large margin in both the angular AUC$@5^{\circ}/10^{\circ}/20^{\circ}$ and the median translation euclidean error and yields an improvement in the median rotation error. As in the previous experiment, our method surpasses SIFT \cite{lowe2004distinctive}, but achieves inferior results compared to  \cite{sun2021loftr,sarlin20superglue} combined with a depth estimation method \cite{ranftl2021vision}.


\noindent \textbf{Across Indoor $\rightarrow$ Outdoor Datasets.} For this experiment we wanted to test our model's ability to generalize from indoor training to outdoor scenes. For that, we test models that were trained on 7Scenes or Scannet on the Cambridge Landmarks dataset~\cite{kendall2015posenet}. Since there are major scale differences between indoor to outdoor scenes, and the GT poses were acquired by SFM, making the positions recovered up to an unknown global scale, we report only angular error for this experiment. Results are shown in Table \ref{tab:cambridge}. When trained on 7Scenes, our method outperforms all other methods by over $25^\circ$ in translation direction and nearly $1^\circ$ better in rotation compared to the second-best method 8Point. Unfortunately, all the methods that were trained on 7Scenes generalize poorly to outdoor. When trained on Scannet our method outperforms RPR [3D-3D] by $6^\circ / 0.3^\circ$ on translation and rotation, respectively. Overall, our method outperforms all other tested methods in this scenario.

\begin{table*}[!ht]
    \centering
    \small
    \begin{tabular}{|l|c|c|c|c|c|c|}
    \hline
    
        Scene/Method & RelPoseNet\cite{laskar2017camera} & 8Point\cite{rockwell20228} & RPR [$q+t$]\cite{arnold2022map} & Ours &RPR [3D-3D]\cite{arnold2022map}-Scannet& Ours-Scannet \\ \hline
        \hline
        KingsCollege & 86.0/18.62 & 90.4/7.889 & 94.8/8.71 &  45.7/6.20 & 31.2/\textbf{4.25}  & \textbf{27.0}/4.39
\\ \hline
        OldHospital & 91.0/18.23 & 87.3/9.674 & 86.9/12.06 & 62.1/10.01 & 36.3/6.31 & \textbf{28.1}/\textbf{5.51}
\\ \hline
        ShopFacade & 88.9/19.36 & 88.9/11.46 & 81.9/14.42 & 37.8/10.27 & 21.6/\textbf{6.68}  &\textbf{17.1}/7.29 
\\ \hline
        StMarysChurch & 92.2/20.15 & 89.9/12.59  & 85.1/14.17 & 51.9/12.09 & 29.8/7.29 & \textbf{21.9}/\textbf{6.25}
\\ \hline
        Average & 89.5/19.09 & 89.1/10.40 &  87.1/15.01       &60.6/9.64 & 29.7/6.13  & \textbf{23.5}/\textbf{5.86} \\ \hline
    \end{tabular}
    \caption{\textbf{Cambridge Landmarks experiment.} The table shows the result of training on 7scenes/Scannet datasets and testing on the Cambridge outdoors dataset. Each entry shows position angular error in degrees (left), and orientation error in degrees (right).}
    \label{tab:cambridge}    
\end{table*}

\noindent\textbf{Limitations}. Pose regression with our method depends on obtaining a reasonable warping of the LoFTR feature maps for the two input images. This may be difficult to achieve when input image pairs are captured under wide baseline conditions or from extreme viewing directions. This appears to be the case in the Map-free dataset, in which many of the image pairs are very partially overlapping and a significant number have no overlap at all, see \cite{arnold2022map}, Fig.~5 therein and the example in Fig.~\ref{fig:mapfree}.
Table~\ref{tab:mapfree} shows the result of training and testing our network on this dataset, compared to various baselines including end-to-end RPR methods and RANSAC-based methods. Our method's results are not as good as theirs and probably are further hurt because our LoFTR encoder is fixed and not trained on this dataset. 

\noindent\textbf{Inference time}. Table \ref{tab:inferenceTime} shows inference time for the different methods. Feature matching models have substantially slower inference times compared to regression-based relative pose models, particularly when combined with a depth prediction model for scale estimation

\begin{table*}[!h]
\centering
\scriptsize
\begin{tabular}{|c|c|c|c|c|c|}
\hline Method & $\begin{array}{l}\text { Precision (VCRE }< \\
90 \mathrm{px} \text { ) \textuparrow}\end{array}$ & $\begin{array}{l}\text { Median Reproj. } \\
\text { Error (px)\textdownarrow }\end{array}$ &  $\begin{array}{l}\text { Precision (Err } <\\
\left.25 \mathrm{~cm}, 5^{\circ}\right)\uparrow\end{array}$ & $\begin{array}{l}\text { Median Trans. } \\
\text { Error }(\mathrm{m})\downarrow\end{array}$ & $\begin{array}{l}\text { Median Rot. } \\
\text { Error }\left({ }^{\circ}\right)\downarrow\end{array}$ \\
\hline LoFTR\cite{sun2021loftr}+5Pt+Depth Scale &  $34.7 \%$ & 167.6  & $15.4 \%$ & 1.98 & 30.5 \\
\hline SG\cite{sarlin20superglue}+5Pt+Depth Scale &  $36.1 \%$ & 160.3 &  $\mathbf{16.8 \%}$ & 1.88 & 25.4 \\
\hline
SIFT\cite{lowe2004distinctive}+5Pt+Depth Scale & $25.0 \%$ & 222.8  & $10.3 \%$ & 2.93 &61.4  \\
\hline
\hline RPR[R(6 D)+t]\cite{arnold2022map} &  $\mathbf{40.2 \%}$ & \textbf{147.1} &  $6.0 \%$ & \textbf{1.68} & \textbf{22.5}  \\
\hline  RPR [3D-3D]\cite{arnold2022map} &  $38.7 \%$ & 148.7 &  $6.0 \%$ & 1.69 & 22.9 \\
\hline RPR disc.\cite{arnold2022map} &$35.4 \%$ & 166.3  & $10.5 \%$ & 1.83 & 23.2  \\
\hline 
Ours  & $35.6 \%$ & 155.3  & $6.9 \%$ & 2.00 & 25.8 \\
\hline
\end{tabular}
\caption{\textbf{Map-Free dataset experiment}. The table shows the result of training and testing on the Map-Free dataset. The depth is predicted by DPT \cite{ranftl2021vision} a monocular depth network, its output is utilized to calculate the scale for image matching-based methods as explained in \cite{arnold2022map}.
Virtual Correspondence Reprojection Error (VCRE) is the average Euclidean reprojection error of virtual 3D-2D correspondences.}
\label{tab:mapfree}
\end{table*}

\begin{table*}[!h]
\centering
\small
    \begin{tabular}{|l|l|c|c|c|c|c|c|}
\hline\multirow{2}{*}{ Experiment } & \multirow{2}{*}{ Method } & \multicolumn{3}{|c|}{ Pose estimation AUC \textuparrow} & \multicolumn{3}{|c|}{ Pose median errors\textdownarrow }\\
\cline{3-8} & & @ $5^{\circ}$ & $@ 10^{\circ}$ & $@ 20^{\circ}$  & trans cm & trans\textdegree & rotation\textdegree\\
\hline
\hline
\multirow{2}{*}{ Backbone } & Dino\cite{caron2021emerging} & 2.0 & 8.2 & 23.7 & 23 & 17.7 & 4.28\\
\cline{2-8}  & ACE\cite{brachmann2023ace}  & 1.7 & 9.1 & 25.6 & 20 &15.6 &4.58\\
\hline Warping & Soft Warping & 3.6  & 14.9 & 35.7 & 16 & 11.6 & 3.5\\
\hline \multirow{6}{*}{ Loss \& Architecture } & Single head $(\q,\tr)$  &3.6  & 15.0 & 35.8 & 16 & 11.4 & 3.5\\
\cline{2-8} & Single head $(\q,\tr)$ w/o cosine similarity  &3.5  & 13.6 & 31.9 & 18 & 13.4 & 3.77\\
\cline{2-8} & Two heads $((\q),(\tr))$  & 3.0  & 12.6 & 32.3  & 17 & 13.0 & 3.27\\
\cline{2-8} & Two heads $(\q,(\tr_{dir},\s))$  & 3.2 & 13.6 & 33.7 & 16 & 12.4 & \textbf{3.07}\\
\cline{2-8} & Three heads $((\q),(\tr_{dir}),(\s))$ & 3.5 & 14.6 & 35.7 &16 &11.7 & 3.11\\
\cline{2-8}  &Single head $(\q,\tr_{dir},\s)$  &  \textbf{3.8} &  \textbf{15.4} & \textbf{36.4} & \textbf{15} & \textbf{11.1} & 3.40 \\
\hline
\end{tabular}
\caption{\textbf{Ablation experiment}. The table shows the result of training on Scannet and testing on the Scannet-1500 test set. With the train and test split specified in the text. Following this ablation we use single-headed architecture with a LoFTR backbone and hard warping.}
\label{tab:ablations_scannet}
\end{table*}

\subsection{Ablations}
We ablate our choice of backbone, disentanglement of the regression heads, and the warping mechanism.

\noindent\textbf{Backbone}.
For the backbone, we compare the use of LoFTR to Dino-ViT~\cite{caron2021emerging} and the ACE encoder~\cite{brachmann2023ace}. Dino-ViT is a self-supervised vision transformer whose features were shown to capture a strong structural and semantic representation \cite{amir2021deep,tumanyan2022splicing}. In our experiment, we concatenated the class token to all patch tokens from the output of the last transformer blocks and pass those to our matching and warping model.
The ACE encoder is a convolutional NN with the same architecture as the DSAC\textsuperscript{*} encoder \cite{turkoglu2021visual}. The encoder is trained on 100 scans from Scannet dataset, and it is attached to regression heads that are trained on individual scenes, using the  DSAC\textsuperscript{*} framework. As \cite{brachmann2023ace} shows, the resulting network serves as a scene-agnostic descriptor for visual relocalization.

\noindent\textbf{Loss and network architecture}.
We compared our proposed loss \eqref{eq:loss} with one in which we drop the cosine similarity term $\mathcal{L}_{\tr_{dir}}(\Ima,\Imb)$. We further added balancing weights and learned them as in \cite{kendall2017geometric}. Additionally, we tested models with two and three heads. Using additional regression heads increases the number of parameters but eliminates balancing weights (since the backbone is pre-trained). Finally, we compared utilizing soft to hard warping in our model. 

We trained and tested our ablated method on the Scannet dataset. Results are shown in Table \ref{tab:ablations_scannet}. The best results are obtained with the LoFTR backbone, hard warping, and a single head with unit balancing weights. Utilizing three heads produced nearly similar results. Replacing LoFTR with either DINO-ViT or ACE yielded significant degradation, justifying our use of LoFTR, an IM model, as the backbone.

\begin{table}[h]
\footnotesize
\centering
\begin{tabular}{ccccc} 
\hline
Time [ms] &LoFTR&SuperGlue&RPR [3D-3D]& Ours\\ 
\hline
Method & 79& 71 & 21 & 26\\
+ Depth estimation & 40& 40 & - & -\\
\hline
Total time & 119& 111 & 21 & 26\\
\hline
\end{tabular}
\caption{\textbf{Inference  time}. Feature matching models are significantly slower than regression-based relative pose models in terms of inference time.}
\label{tab:inferenceTime}
\end{table}

\section{Conclusion}

This paper proposes to use image matching as a pre-training task to learn better feature representations toward a generalizable end-to-end neural network for relative pose regression. Our method relies on a semi-dense feature map produced by a pre-trained LoFTR network, image warping and pose regressor. We showed that this architecture, coupled with a loss function that separates the translation direction and scale, achieves improved results on various datasets and in cross-scene and cross-dataset problems. Our method has effectively reduced the performance gap between RPRs and feature matching models while maintaining significantly faster inference speeds. We also showed limitations in handling image pairs in wide baseline and sharp illumination changes.
We plan to continue exploring network architectures to further improve the accuracy of RPR networks.

\subsection*{Acknowledgments}
Research at the Weizmann Institute was partially supported by the Israel Science Foundation, grant No. 1639/19, by the Israeli Council for Higher Education (CHE) via the Weizmann Data Science Research Center, by the MBZUAI-WIS Joint Program for Artificial Intelligence Research and from the Estates of Tully and Michele Plesser and the Anita James Rosen and Harry Schutzman Foundations.

\newpage
{
    \small
    \bibliographystyle{ieeenat_fullname}
    \bibliography{main}

\begin{thebibliography}{68}
\providecommand{\natexlab}[1]{#1}
\providecommand{\url}[1]{\texttt{#1}}
\expandafter\ifx\csname urlstyle\endcsname\relax
  \providecommand{\doi}[1]{doi: #1}\else
  \providecommand{\doi}{doi: \begingroup \urlstyle{rm}\Url}\fi

\bibitem[Abouelnaga et~al.(2021)Abouelnaga, Bui, and Ilic]{abouelnaga2021distillpose}
Yehya Abouelnaga, Mai Bui, and Slobodan Ilic.
\newblock Distillpose: Lightweight camera localization using auxiliary learning.
\newblock In \emph{2021 IEEE/RSJ International Conference on Intelligent Robots and Systems (IROS)}, pages 7919--7924. IEEE, 2021.

\bibitem[Amir et~al.(2022)Amir, Gandelsman, Bagon, and Dekel]{amir2021deep}
Shir Amir, Yossi Gandelsman, Shai Bagon, and Tali Dekel.
\newblock Deep vit features as dense visual descriptors.
\newblock \emph{ECCVW What is Motion For?}, 2022.

\bibitem[Arnold et~al.(2022)Arnold, Wynn, Vicente, Garcia-Hernando, Monszpart, Prisacariu, Turmukhambetov, and Brachmann]{arnold2022map}
Eduardo Arnold, Jamie Wynn, Sara Vicente, Guillermo Garcia-Hernando, {\'{A}}ron Monszpart, Victor~Adrian Prisacariu, Daniyar Turmukhambetov, and Eric Brachmann.
\newblock Map-free visual relocalization: Metric pose relative to a single image.
\newblock In \emph{ECCV}, 2022.

\bibitem[Balntas et~al.(2018)Balntas, Li, and Prisacariu]{balntas2018relocnet}
Vassileios Balntas, Shuda Li, and Victor Prisacariu.
\newblock Relocnet: Continuous metric learning relocalisation using neural nets.
\newblock In \emph{Proceedings of the European Conference on Computer Vision (ECCV)}, pages 751--767, 2018.

\bibitem[Barath et~al.(2019)Barath, Matas, and Noskova]{barath2019magsac}
Daniel Barath, Jiri Matas, and Jana Noskova.
\newblock Magsac: marginalizing sample consensus.
\newblock In \emph{Proceedings of the IEEE/CVF Conference on Computer Vision and Pattern Recognition}, pages 10197--10205, 2019.

\bibitem[Barath et~al.(2020)Barath, Noskova, Ivashechkin, and Matas]{barath2020magsac++}
Daniel Barath, Jana Noskova, Maksym Ivashechkin, and Jiri Matas.
\newblock Magsac++, a fast, reliable and accurate robust estimator.
\newblock In \emph{Proceedings of the IEEE/CVF conference on computer vision and pattern recognition}, pages 1304--1312, 2020.

\bibitem[Barroso-Laguna et~al.(2023)Barroso-Laguna, Brachmann, Prisacariu, Brostow, and Turmukhambetov]{barroso2023fsnet}
Axel Barroso-Laguna, Eric Brachmann, Victor Prisacariu, Gabriel Brostow, and Daniyar Turmukhambetov.
\newblock Two-view geometry scoring without correspondences.
\newblock In \emph{Proceedings of the IEEE/CVF Conference on Computer Vision and Pattern Recognition}, 2023.

\bibitem[Bay et~al.(2006)Bay, Tuytelaars, and Gool]{bay2006surf}
Herbert Bay, Tinne Tuytelaars, and Luc~Van Gool.
\newblock Surf: Speeded up robust features.
\newblock In \emph{European conference on computer vision}, pages 404--417. Springer, 2006.

\bibitem[Bhowmik et~al.(2020)Bhowmik, Gumhold, Rother, and Brachmann]{bhowmik2020reinforced}
Aritra Bhowmik, Stefan Gumhold, Carsten Rother, and Eric Brachmann.
\newblock Reinforced feature points: Optimizing feature detection and description for a high-level task.
\newblock In \emph{Proceedings of the IEEE/CVF conference on computer vision and pattern recognition}, pages 4948--4957, 2020.

\bibitem[Brachmann and Rother(2019{\natexlab{a}})]{brachmann2019expert}
Eric Brachmann and Carsten Rother.
\newblock Expert sample consensus applied to camera re-localization.
\newblock In \emph{Proceedings of the IEEE/CVF International Conference on Computer Vision}, pages 7525--7534, 2019{\natexlab{a}}.

\bibitem[Brachmann and Rother(2019{\natexlab{b}})]{brachmann2019neural}
Eric Brachmann and Carsten Rother.
\newblock Neural-guided ransac: Learning where to sample model hypotheses.
\newblock In \emph{Proceedings of the IEEE/CVF International Conference on Computer Vision}, pages 4322--4331, 2019{\natexlab{b}}.

\bibitem[Brachmann and Rother(2021)]{brachmann2021visual}
Eric Brachmann and Carsten Rother.
\newblock Visual camera re-localization from rgb and rgb-d images using dsac.
\newblock \emph{IEEE transactions on pattern analysis and machine intelligence}, 44\penalty0 (9):\penalty0 5847--5865, 2021.

\bibitem[Brachmann et~al.(2017)Brachmann, Krull, Nowozin, Shotton, Michel, Gumhold, and Rother]{brachmann2017dsac}
Eric Brachmann, Alexander Krull, Sebastian Nowozin, Jamie Shotton, Frank Michel, Stefan Gumhold, and Carsten Rother.
\newblock Dsac-differentiable ransac for camera localization.
\newblock In \emph{Proceedings of the IEEE conference on computer vision and pattern recognition}, pages 6684--6692, 2017.

\bibitem[Brachmann et~al.(2023)Brachmann, Cavallari, and Prisacariu]{brachmann2023ace}
Eric Brachmann, Tommaso Cavallari, and Victor~Adrian Prisacariu.
\newblock Accelerated coordinate encoding: Learning to relocalize in minutes using rgb and poses.
\newblock In \emph{CVPR}, 2023.

\bibitem[Calonder et~al.(2010)Calonder, Lepetit, Strecha, and Fua]{calonder2010brief}
Michael Calonder, Vincent Lepetit, Christoph Strecha, and Pascal Fua.
\newblock Brief: Binary robust independent elementary features.
\newblock In \emph{European conference on computer vision}, pages 778--792. Springer, 2010.

\bibitem[Caron et~al.(2021)Caron, Touvron, Misra, J\'egou, Mairal, Bojanowski, and Joulin]{caron2021emerging}
Mathilde Caron, Hugo Touvron, Ishan Misra, Herv\'e J\'egou, Julien Mairal, Piotr Bojanowski, and Armand Joulin.
\newblock Emerging properties in self-supervised vision transformers.
\newblock In \emph{Proceedings of the International Conference on Computer Vision (ICCV)}, 2021.

\bibitem[Choy et~al.(2020)Choy, Lee, Ranftl, Park, and Koltun]{choy2020high}
Christopher Choy, Junha Lee, Ren{\'e} Ranftl, Jaesik Park, and Vladlen Koltun.
\newblock High-dimensional convolutional networks for geometric pattern recognition.
\newblock In \emph{Proceedings of the IEEE/CVF conference on computer vision and pattern recognition}, pages 11227--11236, 2020.

\bibitem[Dai et~al.(2017)Dai, Chang, Savva, Halber, Funkhouser, and Nie{\ss}ner]{dai2017scannet}
Angela Dai, Angel~X Chang, Manolis Savva, Maciej Halber, Thomas Funkhouser, and Matthias Nie{\ss}ner.
\newblock Scannet: Richly-annotated 3d reconstructions of indoor scenes.
\newblock In \emph{Proceedings of the IEEE conference on computer vision and pattern recognition}, pages 5828--5839, 2017.

\bibitem[DeTone et~al.(2018)DeTone, Malisiewicz, and Rabinovich]{detone2018superpoint}
Daniel DeTone, Tomasz Malisiewicz, and Andrew Rabinovich.
\newblock Superpoint: Self-supervised interest point detection and description.
\newblock In \emph{Proceedings of the IEEE conference on computer vision and pattern recognition workshops}, pages 224--236, 2018.

\bibitem[Dosovitskiy et~al.(2020)Dosovitskiy, Beyer, Kolesnikov, Weissenborn, Zhai, Unterthiner, Dehghani, Minderer, Heigold, Gelly, et~al.]{dosovitskiy2020image}
Alexey Dosovitskiy, Lucas Beyer, Alexander Kolesnikov, Dirk Weissenborn, Xiaohua Zhai, Thomas Unterthiner, Mostafa Dehghani, Matthias Minderer, Georg Heigold, Sylvain Gelly, et~al.
\newblock An image is worth 16x16 words: Transformers for image recognition at scale.
\newblock \emph{arXiv preprint arXiv:2010.11929}, 2020.

\bibitem[Dusmanu et~al.(2019)Dusmanu, Rocco, Pajdla, Pollefeys, Sivic, Torii, and Sattler]{dusmanu2019d2}
Mihai Dusmanu, Ignacio Rocco, Tomas Pajdla, Marc Pollefeys, Josef Sivic, Akihiko Torii, and Torsten Sattler.
\newblock D2-net: A trainable cnn for joint description and detection of local features.
\newblock In \emph{Proceedings of the ieee/cvf conference on computer vision and pattern recognition}, pages 8092--8101, 2019.

\bibitem[En et~al.(2018)En, Lechervy, and Jurie]{en2018rpnet}
Sovann En, Alexis Lechervy, and Fr{\'e}d{\'e}ric Jurie.
\newblock Rpnet: An end-to-end network for relative camera pose estimation.
\newblock In \emph{Proceedings of the European Conference on Computer Vision (ECCV) Workshops}, pages 0--0, 2018.

\bibitem[Fischler and Bolles(1981)]{fischler1981random}
Martin~A Fischler and Robert~C Bolles.
\newblock Random sample consensus: a paradigm for model fitting with applications to image analysis and automated cartography.
\newblock \emph{Communications of the ACM}, 24\penalty0 (6):\penalty0 381--395, 1981.

\bibitem[Glocker et~al.(2013)Glocker, Izadi, Shotton, and Criminisi]{glocker2013real}
Ben Glocker, Shahram Izadi, Jamie Shotton, and Antonio Criminisi.
\newblock Real-time rgb-d camera relocalization.
\newblock In \emph{2013 IEEE International Symposium on Mixed and Augmented Reality (ISMAR)}, pages 173--179. IEEE, 2013.

\bibitem[Hartley and Zisserman(2003)]{hartley2003multiple}
Richard Hartley and Andrew Zisserman.
\newblock \emph{Multiple view geometry in computer vision}.
\newblock Cambridge university press, 2003.

\bibitem[Hartley(1997)]{hartley1997defense}
Richard~I Hartley.
\newblock In defense of the eight-point algorithm.
\newblock \emph{IEEE Transactions on pattern analysis and machine intelligence}, 19\penalty0 (6):\penalty0 580--593, 1997.

\bibitem[He et~al.(2016)He, Zhang, Ren, and Sun]{he2016deep}
Kaiming He, Xiangyu Zhang, Shaoqing Ren, and Jian Sun.
\newblock Deep residual learning for image recognition.
\newblock In \emph{Proceedings of the IEEE conference on computer vision and pattern recognition}, pages 770--778, 2016.

\bibitem[Jiang et~al.(2021)Jiang, Trulls, Hosang, Tagliasacchi, and Yi]{jiang2021cotr}
Wei Jiang, Eduard Trulls, Jan Hosang, Andrea Tagliasacchi, and Kwang~Moo Yi.
\newblock {COTR}: Correspondence transformer for matching across images.
\newblock In \emph{Proceedings of the IEEE/CVF International Conference on Computer Vision}, pages 6207--6217, 2021.

\bibitem[Kendall and Cipolla(2017)]{kendall2017geometric}
Alex Kendall and Roberto Cipolla.
\newblock Geometric loss functions for camera pose regression with deep learning.
\newblock In \emph{Proceedings of the IEEE conference on computer vision and pattern recognition}, pages 5974--5983, 2017.

\bibitem[Kendall et~al.(2015)Kendall, Grimes, and Cipolla]{kendall2015posenet}
Alex Kendall, Matthew Grimes, and Roberto Cipolla.
\newblock Posenet: A convolutional network for real-time 6-dof camera relocalization.
\newblock In \emph{Proceedings of the IEEE international conference on computer vision}, pages 2938--2946, 2015.

\bibitem[Laskar et~al.(2017)Laskar, Melekhov, Kalia, and Kannala]{laskar2017camera}
Zakaria Laskar, Iaroslav Melekhov, Surya Kalia, and Juho Kannala.
\newblock Camera relocalization by computing pairwise relative poses using convolutional neural network.
\newblock In \emph{Proceedings of the IEEE International Conference on Computer Vision Workshops}, pages 929--938, 2017.

\bibitem[Lin et~al.(2023)Lin, Zhang, Ramanan, and Tulsiani]{lin2023relpose++}
Amy Lin, Jason~Y Zhang, Deva Ramanan, and Shubham Tulsiani.
\newblock Relpose++: Recovering 6d poses from sparse-view observations.
\newblock \emph{arXiv preprint arXiv:2305.04926}, 2023.

\bibitem[Lin et~al.(2017)Lin, Doll{\'a}r, Girshick, He, Hariharan, and Belongie]{lin2017feature}
Tsung-Yi Lin, Piotr Doll{\'a}r, Ross Girshick, Kaiming He, Bharath Hariharan, and Serge Belongie.
\newblock Feature pyramid networks for object detection.
\newblock In \emph{Proceedings of the IEEE conference on computer vision and pattern recognition}, pages 2117--2125, 2017.

\bibitem[Lindenberger et~al.(2023)Lindenberger, Sarlin, and Pollefeys]{lindenberger2023lightglue}
Philipp Lindenberger, Paul-Edouard Sarlin, and Marc Pollefeys.
\newblock {LightGlue: Local Feature Matching at Light Speed}.
\newblock In \emph{ICCV}, 2023.

\bibitem[Liu et~al.(2019)Liu, Kim, Gu, Furukawa, and Kautz]{liu2019planercnn}
Chen Liu, Kihwan Kim, Jinwei Gu, Yasutaka Furukawa, and Jan Kautz.
\newblock Planercnn: 3d plane detection and reconstruction from a single image.
\newblock In \emph{Proceedings of the IEEE/CVF Conference on Computer Vision and Pattern Recognition}, pages 4450--4459, 2019.

\bibitem[Lowe(2004)]{lowe2004distinctive}
David~G Lowe.
\newblock Distinctive image features from scale-invariant keypoints.
\newblock \emph{International journal of computer vision}, 60\penalty0 (2):\penalty0 91--110, 2004.

\bibitem[Melekhov et~al.(2017)Melekhov, Ylioinas, Kannala, and Rahtu]{melekhov2017relative}
Iaroslav Melekhov, Juha Ylioinas, Juho Kannala, and Esa Rahtu.
\newblock Relative camera pose estimation using convolutional neural networks.
\newblock In \emph{International Conference on Advanced Concepts for Intelligent Vision Systems}, pages 675--687. Springer, 2017.

\bibitem[Nist{\'e}r(2004)]{nister2004efficient}
David Nist{\'e}r.
\newblock An efficient solution to the five-point relative pose problem.
\newblock \emph{IEEE transactions on pattern analysis and machine intelligence}, 26\penalty0 (6):\penalty0 756--770, 2004.

\bibitem[{\"O}zye{\c{s}}il et~al.(2017){\"O}zye{\c{s}}il, Voroninski, Basri, and Singer]{ozyecsil2017survey}
Onur {\"O}zye{\c{s}}il, Vladislav Voroninski, Ronen Basri, and Amit Singer.
\newblock A survey of structure from motion*.
\newblock \emph{Acta Numerica}, 26:\penalty0 305--364, 2017.

\bibitem[Paszke et~al.(2019)Paszke, Gross, Massa, Lerer, Bradbury, Chanan, Killeen, Lin, Gimelshein, Antiga, et~al.]{paszke2019pytorch}
Adam Paszke, Sam Gross, Francisco Massa, Adam Lerer, James Bradbury, Gregory Chanan, Trevor Killeen, Zeming Lin, Natalia Gimelshein, Luca Antiga, et~al.
\newblock Pytorch: An imperative style, high-performance deep learning library.
\newblock \emph{Advances in neural information processing systems}, 32, 2019.

\bibitem[Raguram et~al.(2008)Raguram, Frahm, and Pollefeys]{raguram2008comparative}
Rahul Raguram, Jan-Michael Frahm, and Marc Pollefeys.
\newblock A comparative analysis of ransac techniques leading to adaptive real-time random sample consensus.
\newblock In \emph{European conference on computer vision}, pages 500--513. Springer, 2008.

\bibitem[Ranftl and Koltun(2018)]{ranftl2018deep}
Ren{\'e} Ranftl and Vladlen Koltun.
\newblock Deep fundamental matrix estimation.
\newblock In \emph{Proceedings of the European conference on computer vision (ECCV)}, pages 284--299, 2018.

\bibitem[Ranftl et~al.(2021)Ranftl, Bochkovskiy, and Koltun]{ranftl2021vision}
Ren{\'e} Ranftl, Alexey Bochkovskiy, and Vladlen Koltun.
\newblock Vision transformers for dense prediction.
\newblock In \emph{Proceedings of the IEEE/CVF international conference on computer vision}, pages 12179--12188, 2021.

\bibitem[Reizenstein et~al.(2021)Reizenstein, Shapovalov, Henzler, Sbordone, Labatut, and Novotny]{reizenstein2021common}
Jeremy Reizenstein, Roman Shapovalov, Philipp Henzler, Luca Sbordone, Patrick Labatut, and David Novotny.
\newblock Common objects in 3d: Large-scale learning and evaluation of real-life 3d category reconstruction.
\newblock In \emph{Proceedings of the IEEE/CVF International Conference on Computer Vision}, pages 10901--10911, 2021.

\bibitem[Revaud et~al.(2019)Revaud, De~Souza, Humenberger, and Weinzaepfel]{revaud2019r2d2}
Jerome Revaud, Cesar De~Souza, Martin Humenberger, and Philippe Weinzaepfel.
\newblock R2d2: Reliable and repeatable detector and descriptor.
\newblock \emph{Advances in neural information processing systems}, 32, 2019.

\bibitem[Riba et~al.(2020)Riba, Mishkin, Ponsa, Rublee, and Bradski]{riba2020kornia}
Edgar Riba, Dmytro Mishkin, Daniel Ponsa, Ethan Rublee, and Gary Bradski.
\newblock Kornia: an open source differentiable computer vision library for pytorch.
\newblock In \emph{Proceedings of the IEEE/CVF Winter Conference on Applications of Computer Vision}, pages 3674--3683, 2020.

\bibitem[Rockwell et~al.(2022)Rockwell, Johnson, and Fouhey]{rockwell20228}
Chris Rockwell, Justin Johnson, and David~F Fouhey.
\newblock The 8-point algorithm as an inductive bias for relative pose prediction by {ViTs}.
\newblock \emph{arXiv preprint arXiv:2208.08988}, 2022.

\bibitem[Rosten and Drummond(2006)]{rosten2006machine}
Edward Rosten and Tom Drummond.
\newblock Machine learning for high-speed corner detection.
\newblock In \emph{European conference on computer vision}, pages 430--443. Springer, 2006.

\bibitem[Rublee et~al.(2011)Rublee, Rabaud, Konolige, and Bradski]{rublee2011orb}
Ethan Rublee, Vincent Rabaud, Kurt Konolige, and Gary Bradski.
\newblock Orb: An efficient alternative to sift or surf.
\newblock In \emph{2011 International conference on computer vision}, pages 2564--2571. Ieee, 2011.

\bibitem[Sarlin et~al.(2020)Sarlin, DeTone, Malisiewicz, and Rabinovich]{sarlin20superglue}
Paul-Edouard Sarlin, Daniel DeTone, Tomasz Malisiewicz, and Andrew Rabinovich.
\newblock {SuperGlue}: Learning feature matching with graph neural networks.
\newblock In \emph{CVPR}, 2020.

\bibitem[Sch\"{o}nberger and Frahm(2016)]{schoenberger2016sfm}
Johannes~Lutz Sch\"{o}nberger and Jan-Michael Frahm.
\newblock Structure-from-motion revisited.
\newblock In \emph{Conference on Computer Vision and Pattern Recognition (CVPR)}, 2016.

\bibitem[Snavely et~al.(2006)Snavely, Seitz, and Szeliski]{snavely2006photo}
Noah Snavely, Steven~M Seitz, and Richard Szeliski.
\newblock Photo tourism: exploring photo collections in 3d.
\newblock In \emph{ACM siggraph 2006 papers}, pages 835--846. 2006.

\bibitem[Sun et~al.(2021)Sun, Shen, Wang, Bao, and Zhou]{sun2021loftr}
Jiaming Sun, Zehong Shen, Yuang Wang, Hujun Bao, and Xiaowei Zhou.
\newblock {LoFTR}: Detector-free local feature matching with transformers.
\newblock \emph{CVPR}, 2021.

\bibitem[Torii et~al.(2015)Torii, Arandjelovic, Sivic, Okutomi, and Pajdla]{torii201524}
Akihiko Torii, Relja Arandjelovic, Josef Sivic, Masatoshi Okutomi, and Tomas Pajdla.
\newblock 24/7 place recognition by view synthesis.
\newblock In \emph{Proceedings of the IEEE conference on computer vision and pattern recognition}, pages 1808--1817, 2015.

\bibitem[Truong et~al.(2020{\natexlab{a}})Truong, Danelljan, Gool, and Timofte]{truong2020gocor}
Prune Truong, Martin Danelljan, Luc~V Gool, and Radu Timofte.
\newblock Gocor: Bringing globally optimized correspondence volumes into your neural network.
\newblock \emph{Advances in Neural Information Processing Systems}, 33:\penalty0 14278--14290, 2020{\natexlab{a}}.

\bibitem[Truong et~al.(2020{\natexlab{b}})Truong, Danelljan, and Timofte]{truong2020glu}
Prune Truong, Martin Danelljan, and Radu Timofte.
\newblock Glu-net: Global-local universal network for dense flow and correspondences.
\newblock In \emph{Proceedings of the IEEE/CVF conference on computer vision and pattern recognition}, pages 6258--6268, 2020{\natexlab{b}}.

\bibitem[Truong et~al.(2021)Truong, Danelljan, Van~Gool, and Timofte]{truong2021learning}
Prune Truong, Martin Danelljan, Luc Van~Gool, and Radu Timofte.
\newblock Learning accurate dense correspondences and when to trust them.
\newblock In \emph{Proceedings of the IEEE/CVF Conference on Computer Vision and Pattern Recognition}, pages 5714--5724, 2021.

\bibitem[Tumanyan et~al.(2022)Tumanyan, Bar-Tal, Bagon, and Dekel]{tumanyan2022splicing}
Narek Tumanyan, Omer Bar-Tal, Shai Bagon, and Tali Dekel.
\newblock Splicing vit features for semantic appearance transfer.
\newblock In \emph{Proceedings of the IEEE/CVF Conference on Computer Vision and Pattern Recognition}, pages 10748--10757, 2022.

\bibitem[Turkoglu et~al.(2021)Turkoglu, Brachmann, Schindler, Brostow, and Monszpart]{turkoglu2021visual}
Mehmet~Ozgur Turkoglu, Eric Brachmann, Konrad Schindler, Gabriel~J Brostow, and Aron Monszpart.
\newblock Visual camera re-localization using graph neural networks and relative pose supervision.
\newblock In \emph{2021 International Conference on 3D Vision (3DV)}, pages 145--155. IEEE, 2021.

\bibitem[Tyszkiewicz et~al.(2020)Tyszkiewicz, Fua, and Trulls]{tyszkiewicz2020disk}
Micha{\l} Tyszkiewicz, Pascal Fua, and Eduard Trulls.
\newblock Disk: Learning local features with policy gradient.
\newblock \emph{Advances in Neural Information Processing Systems}, 33:\penalty0 14254--14265, 2020.

\bibitem[Vaswani et~al.(2017)Vaswani, Shazeer, Parmar, Uszkoreit, Jones, Gomez, Kaiser, and Polosukhin]{vaswani2017attention}
Ashish Vaswani, Noam Shazeer, Niki Parmar, Jakob Uszkoreit, Llion Jones, Aidan~N Gomez, {\L}ukasz Kaiser, and Illia Polosukhin.
\newblock Attention is all you need.
\newblock \emph{Advances in neural information processing systems}, 30, 2017.

\bibitem[Winkelbauer et~al.(2021)Winkelbauer, Denninger, and Triebel]{winkelbauer2021learning}
Dominik Winkelbauer, Maximilian Denninger, and Rudolph Triebel.
\newblock Learning to localize in new environments from synthetic training data.
\newblock In \emph{2021 IEEE International Conference on Robotics and Automation (ICRA)}, pages 5840--5846. IEEE, 2021.

\bibitem[Wu(2011)]{wu2011visualsfm}
Changchang Wu.
\newblock Visualsfm: A visual structure from motion system.
\newblock \emph{http://www. cs. washington. edu/homes/ccwu/vsfm}, 2011.

\bibitem[Wu(2013)]{wu2013towards}
Changchang Wu.
\newblock Towards linear-time incremental structure from motion.
\newblock In \emph{2013 International Conference on 3D Vision-3DV 2013}, pages 127--134. IEEE, 2013.

\bibitem[Yi et~al.(2016)Yi, Trulls, Lepetit, and Fua]{yi2016lift}
Kwang~Moo Yi, Eduard Trulls, Vincent Lepetit, and Pascal Fua.
\newblock Lift: Learned invariant feature transform.
\newblock In \emph{European conference on computer vision}, pages 467--483. Springer, 2016.

\bibitem[Yi et~al.(2018)Yi, Trulls, Ono, Lepetit, Salzmann, and Fua]{yi2018learning}
Kwang~Moo Yi, Eduard Trulls, Yuki Ono, Vincent Lepetit, Mathieu Salzmann, and Pascal Fua.
\newblock Learning to find good correspondences.
\newblock In \emph{Proceedings of the IEEE conference on computer vision and pattern recognition}, pages 2666--2674, 2018.

\bibitem[Zhang et~al.(2022)Zhang, Ramanan, and Tulsiani]{zhang2022relpose}
Jason~Y Zhang, Deva Ramanan, and Shubham Tulsiani.
\newblock Relpose: Predicting probabilistic relative rotation for single objects in the wild.
\newblock In \emph{European Conference on Computer Vision}, pages 592--611. Springer, 2022.

\bibitem[Zhou et~al.(2020)Zhou, Sattler, Pollefeys, and Leal-Taixe]{zhou2020learn}
Qunjie Zhou, Torsten Sattler, Marc Pollefeys, and Laura Leal-Taixe.
\newblock To learn or not to learn: Visual localization from essential matrices.
\newblock In \emph{2020 IEEE International Conference on Robotics and Automation (ICRA)}, pages 3319--3326. IEEE, 2020.

\end{thebibliography}
}

\clearpage
\section*{Appendix}
\renewcommand\thesection{\Alph{section}}
\setcounter{section}{0}
\section{Implementation Details}

Our method is implemented using Pytorch~\cite{paszke2019pytorch} and Kornia~\cite{riba2020kornia}. All models are trained and evaluated using the same strategy.
In all the experiments, we use the Adam optimizer initialized with the default values $\beta_1 =
0.9$ and $\beta_2 = 0.999$. The models are trained respectively for 80/60 epochs on the 7Scenes/Scannet datasets with an effective batch size of 200. We use a linear decay strategy, where the learning rate is reduced by a factor of 0.9 every 6 epochs, starting with a learning rate of  $1 \times 10^{-3}$. Using four NVIDIA Tesla A40 GPUs, the training time of our model on the full 7Scenes dataset is roughly 24 hours and on Scannet roughly 72 hours. Using Nvidia V100 GPU the inference time of our model is 26ms. 
We use early stopping with the validation dataset for all methods to determine the best set of weights.

\section{Network Architecture}
We reduced the size of the images to size of \(256\times341 \) following \cite{laskar2017camera}. Therefore, the feature pyramid network \cite{yi2018learning} in LoFTR\cite{sun2021loftr} outputs a \(256\times32\times42\) feature map per image. The resolution is preserved through the attention layers.
The warping module concatenates the feature maps of matching points in the two images and adds $5$ channels, including the coordinates of the two matching points and the matching score. This results in a 2D feature map as output of size \(517\times32\times42\). Which we feed then to a ResNet block, which results in a feature vector of size 1152 (after flattening it). This feature vector we then feed into a three different MLPs which output the predicted translation direction, scale and the orientation.

\section{Ground Truth Relative Pose in the 7Scenes Dataset }
We note that the relative camera positions in the 7scenes \cite{glocker2013real} dataset, obtained from RelPoseNet\cite{laskar2017camera}, are erroneously set to $\tr_i-\tr_j$ where $\tr_i,\tr_j$ represent the absolute position of the $i^\mathrm{th}$ and  $j^\mathrm{th}$ cameras, instead of $\tr_i-R_{ij}^T\tr_j$ as it should be, and how it's computed in the other datasets, where $R_{ij}$ denotes the relative orientation between cameras $i$ and $j$. We rectify this error in our experiments, when we train the different relative pose regression methods on the 7scenes dataset.

\begin{figure*}
    \centering
    \includegraphics[width=\textwidth]{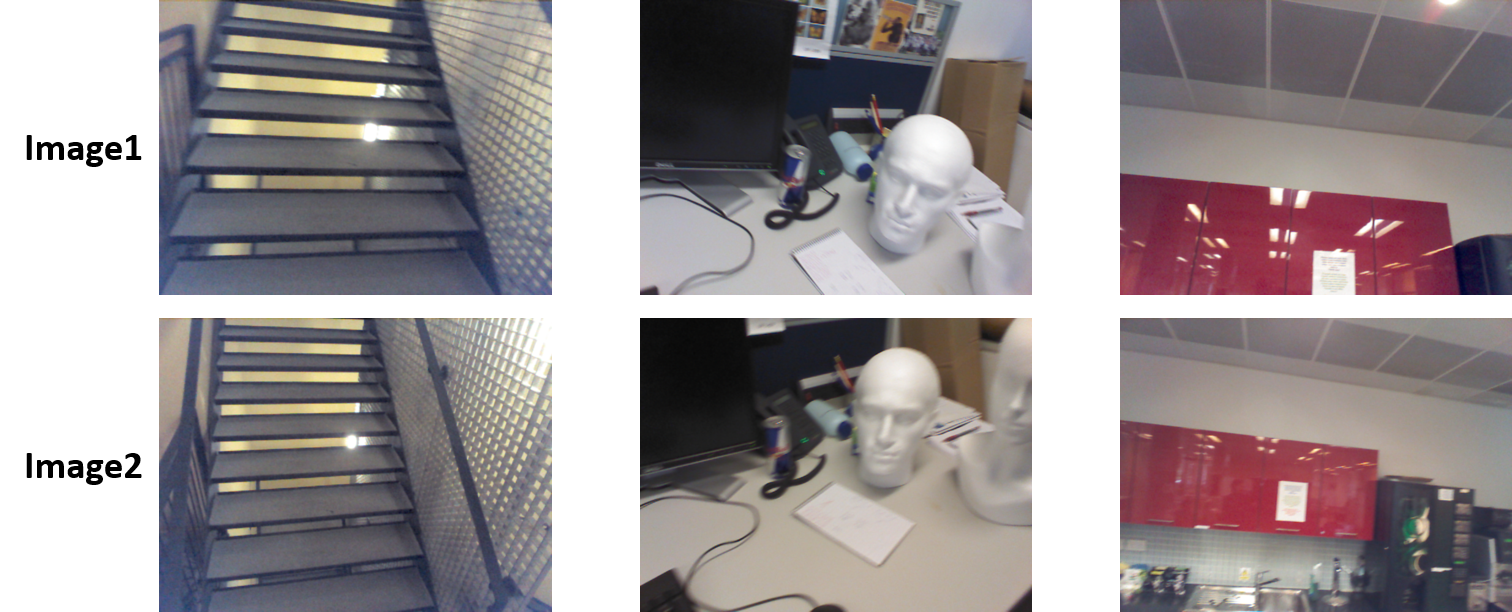}
    \caption{\textbf{7scenes dataset.} Pairs of images from the 7Scenes dataset, including scenes with repetitive patterns.} \label{fig:dataExamples}
\end{figure*}

\begin{figure*}
    \centering
    \includegraphics[width=\textwidth]{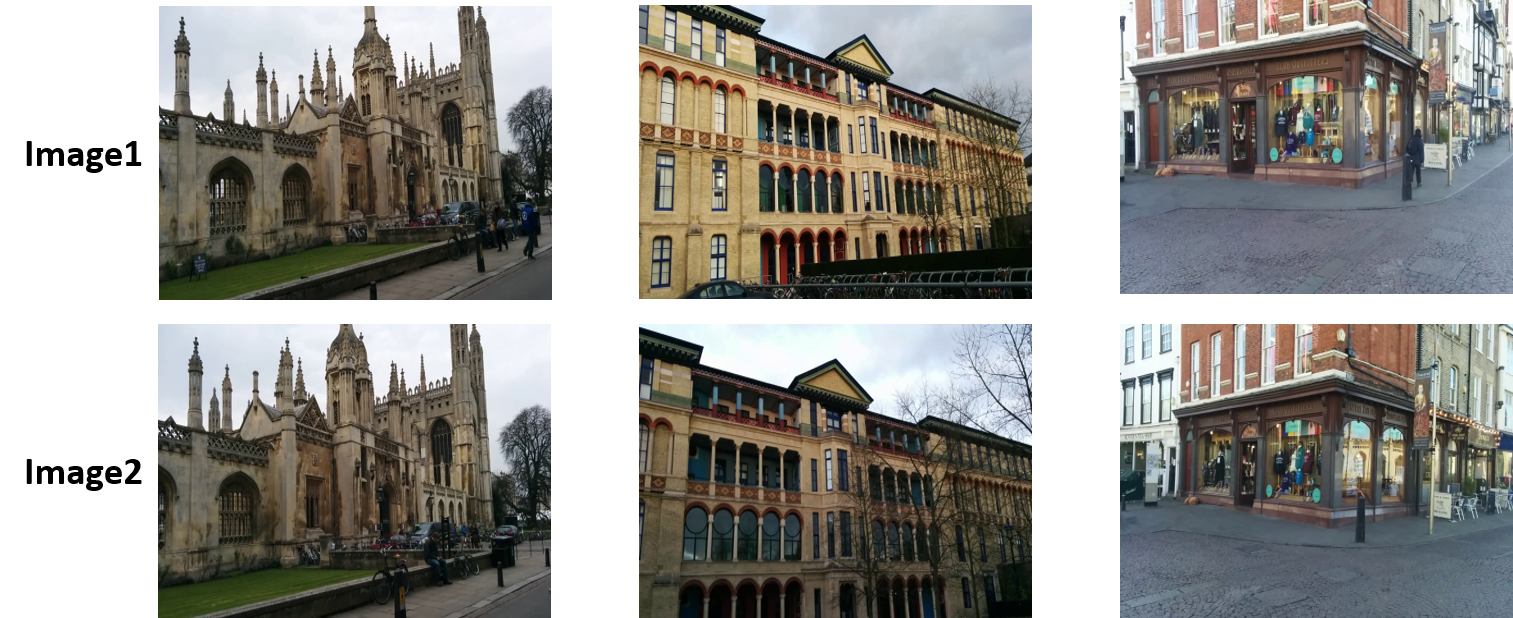}
    \caption{\textbf{Cambridge Landmarks dataset.} Pairs of images from the Kings College, Old Hospital and Shop Facade from the Cambridge Landmarks dataset. Notice the different lighting conditions.} \label{fig:dataExamples}
\end{figure*}
\begin{figure*}
    \centering
    \includegraphics[width=\textwidth]{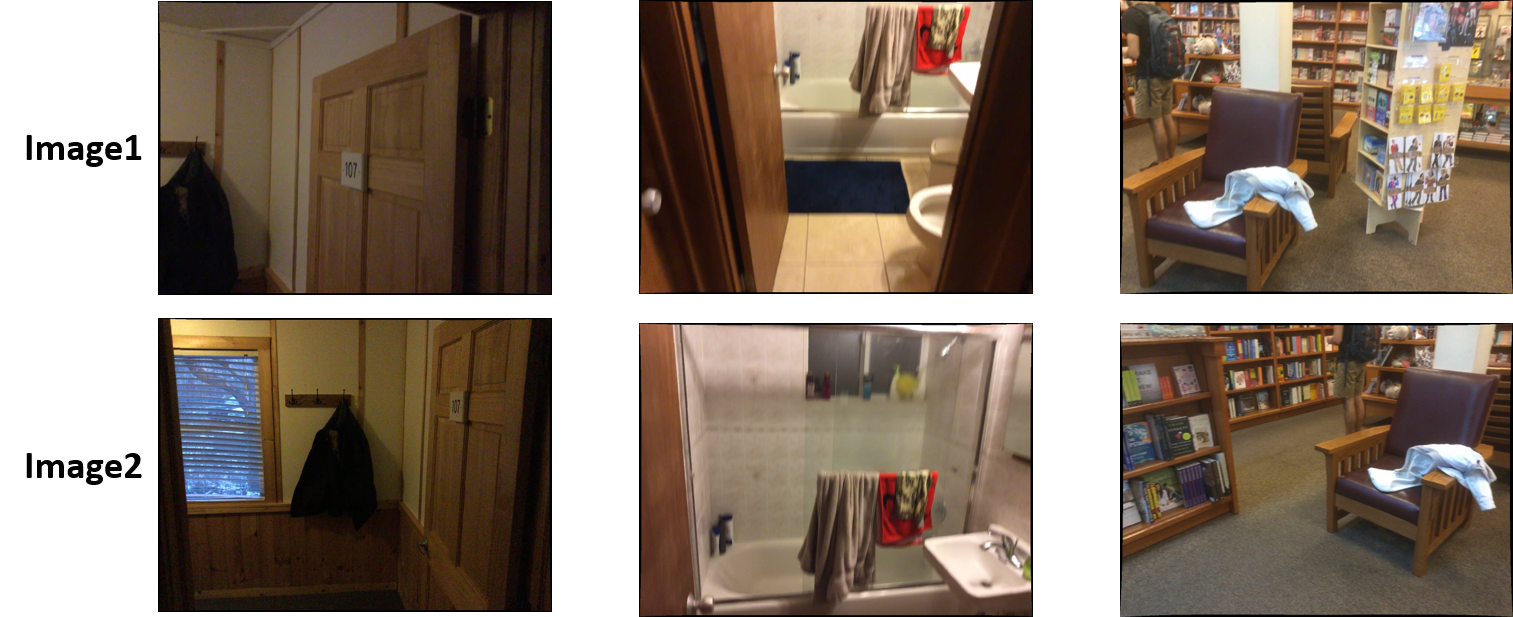}
    \caption{\textbf{Scannet dataset.} Pairs of images from the Scannet dataset, including scenes with repetitive patterns.} \label{fig:dataExamples}
\end{figure*}

\end{document}